# Less is More: A Stealthy and Efficient Adversarial Attack Method for DRL-based Autonomous Driving Policies

Junchao Fan, Xuyang Lei, Xiaolin Chang, *Senior Member*, *IEEE*, Jelena Mišić, *Fellow*, *IEEE*, Vojislav B. Mišić, *Senior Member*, *IEEE*

*Abstract*—Despite significant advancements in deep reinforcement learning (DRL)-based autonomous driving policies, these policies still exhibit vulnerability to adversarial attacks. This vulnerability poses a formidable challenge to the practical deployment of these policies in autonomous driving. Designing effective adversarial attacks is an indispensable prerequisite for enhancing the robustness of these policies. In view of this, we present a novel stealthy and efficient adversarial attack method for DRL-based autonomous driving policies. Specifically, we introduce a DRL-based adversary designed to trigger safety violations (e.g., collisions) by injecting adversarial samples at critical moments. We model the attack as a mixed-integer optimization problem and formulate it as a Markov decision process. Then, we train the adversary to learn the optimal policy for attacking at critical moments without domain knowledge. Furthermore, we introduce attack-related information and a trajectory clipping method to enhance the learning capability of the adversary. Finally, we validate our method in an unprotected left-turn scenario across different traffic densities. The experimental results show that our method achieves more than 90% collision rate within three attacks in most cases. Furthermore, our method achieves more than 130% improvement in attack efficiency compared to the unlimited attack method.

*Index Terms*—Autonomous vehicle, adversarial attacks, deep reinforcement learning, robust decision making

## I. INTRODUCTION

IN recent years, autonomous driving has achieved remarkable progress, propelled by rapid advancements in sensors, cameras, radar, and artificial intelligence (AI) [1][2]. As a prominent paradigm in AI, deep reinforcement learning (DRL) has been widely applied to develop autonomous driving policies, owing to its remarkable capability to address complex decision-making problems [3]. Current DRL-based autonomous driving strategies have demonstrated competitive performance across various tasks, such as highway driving [4]-[6], on-ramp merging [7]-[9], and intersection navigation [3][10][11]. However, the vulnerability of DRL to adversarial attacks has raised concerns about the practical application of such autonomous driving models [12][13]. It is crucial to understand the potential and cost of adversarial attacks against DRL-based autonomous driving policies, as this knowledge can inform the design of more robust and secure algorithms for trustworthy autonomous driving.

### A. Motivation

Some studies have proposed some adversarial attack methods against DRL-based autonomous driving policies and proved that these policies are not robust [13]-[15]. However, despite significant progress, existing approaches still face two key challenges.

1) **Not worst-case oriented**: Current studies mostly focus on either maximizing policy deviation or minimizing agent's rewards by adding adversarial perturbations [16]. These attacks may only result in sudden acceleration or braking, leading to relatively minor effects, such as reduced ride comfort or traffic efficiency. However, they fail to address critical safety challenges in trustworthy autonomous driving, such as safety violations like red light running or collisions.

2) **Fail to exploit critical moments**: Most studies perturb the agent's observation at each time step throughout the entire episode. However, critical moments that could lead to safety violations in autonomous driving are rare during the driving process. Taking an unprotected left-turn task as an example, critical moments with a high risk of safety violations are typically confined to the few seconds during which the ego vehicle crosses the intersection. Optimizing autonomous driving policies under current attack methods may fail to effectively enhance robustness. Moreover, current attack methods are not stealthy enough to avoid detection by the agent due to their high attack frequency.

The key to addressing the above challenges lies in enhancing the intentionality, efficiency, and stealthy of the attacks. Some explorations have been conducted in the field of DRL to improve the efficiency and stealthy of attacks. Lin *et al.* [17] proposed a strategically-timed attack method that triggers an attack when the value of the relative action preference function (RAPF) exceeds a pre-calculated threshold. Li *et al.* [18] further considered the differences in the probabilities of all actions based on RAPF, making the judgment of critical moments more precise. Sun *et al.* [19] proposed a critical point attack in which the adversary constructs a domain-specific model to predict subsequent states and assess the damage of each potential attack policy. A damage awareness metric is used as a threshold to determine critical moments. However, the above attacks require careful design and manual tuning of domain-specific metrics or

J. Fan, X. Lei, and X. Chang are with Beijing Key Laboratory of Security and Privacy in Intelligent Transportation, Beijing Jiaotong University, China. E-mail: {23111144, 23125353, xlchang}@bjtu.edu.cn.

J. Mišić and V. B. Mišić are with Toronto Metropolitan University, Toronto, ON, Canada. E-mail: {jmisic, vmisic} @torontomu.ca.

thresholds, limiting their efficiency and scalability to more complex scenarios. As a result, some studies have attempted to leverage the potential of DRL to create an adversary capable of both selecting critical moments and executing attack actions. Lin *et al.* [17] proposed an antagonist attack in which a DRL-based adversary is trained to learn both when-to-attack and how-to-attack policies, guided by the reward function of the victim agent. However, this approach overlooks the sparsity of perturbation injection, which may lead to suboptimal attack policies. Mo *et al.* [20] proposed a decoupled adversarial policy in which trajectory clipping and padding are used to mitigate the impact of sparse perturbation injection. Moreover, a pre-constructed database is used to provide universal perturbations to induce the victim agent to choose adversarial actions with 100% probability. However, designing universal perturbations for autonomous driving tasks with continuous action space and complex state space is challenging. Since existing research cannot be effectively applied, it is crucial to explore new adversarial attacks in autonomous driving scenarios to address the above challenges. We believe this is a crucial direction for further enhancing the robustness of DRL-based autonomous driving policies, yet it has not been fully explored.

*B. Contribution*

In this paper, we propose a novel stealthy and efficient adversarial attack method, aiming to induce safety violations with the minimum number of attacks. To further improve the stealthy of the attack, the perturbations imposed by each attack are bounded. In our method, we first model the attack as a mixed integer optimization problem which is difficult to solve. Moreover, the problem becomes even more challenging to solve as the state space grows exponentially with the increase of time steps. Therefore, we model this problem as a Markov decision process (MDP) and introduce an adversary based on the proximal policy optimization algorithm (PPO) [21] to solve the MDP to learn the optimal attack policy. We design a reward function to guide the adversary to induce the ego vehicle to perform safety-violating behaviors, rather than simply minimizing the reward of the ego vehicle. Different from existing research where the adversary and victim share the same state space, we further introduce the remaining number of attacks and the ego vehicle's origin action without perturbation into the adversary's state space. Additionally, a trajectory clipping method is employed to optimize the adversary's training. We conduct our experiments in an unprotected left turn scenario with different traffic densities based on SUMO [22]. Experiments show that our method can achieve collision rates of over 90%, with an average of only three attack steps in most cases. Moreover, compared to the attack method without the attack frequency constraint, our method improves the attack efficiency by at least 130% in all cases. The contributions of our work are summarized as follows.

- We propose a novel stealthy and efficient adversarial attack method against DRL-based autonomous driving policies. This method aims to minimize the number of attacks while maximizing safety violations. To further optimize training, we incorporate additional attack-related information into the adversary's observations to enhance its capabilities and use a trajectory clipping method to improve sample quality.
- We conduct extensive experiments on various DRL-based autonomous driving policies, demonstrating the superiority of our method compared to existing attack methods. In addition, we propose a new evaluation metric, attack efficiency, offering a new perspective for evaluating the effectiveness of adversarial attacks against DRL-based autonomous driving policies. Finally, we validate the contributions of each component through ablation studies.

The remaining paper is structured as follows. Section II introduces the related work. Section III provides the details on the preliminaries on MDPs and the problem formulation. In Section IV, we present our attack method. The experimental results are discussed and analyzed in Section V, followed by the conclusion in Section VI.

## II. RELATED WORK

This section first reviews the recent research on DRL-based autonomous driving policies in Section II.A, followed by a review of adversarial attack methods targeting these policies in Section II.B.

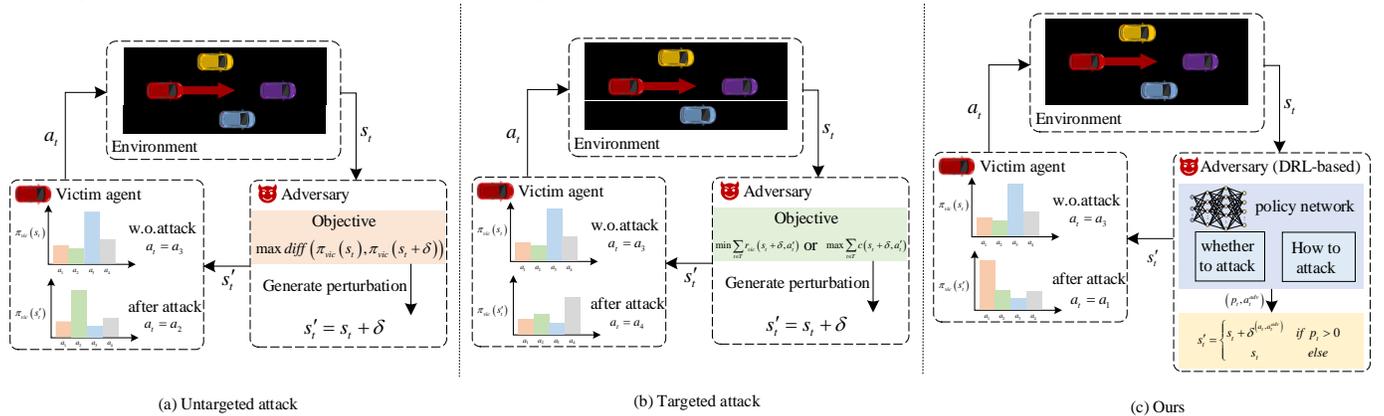

**Fig. 1.** The difference between the existing attack paradigms and our attack method.

*A. DRL-based Autonomous Driving*

The rapid advancements in machine intelligence and data-driven approaches have propelled DRL to the forefront of autonomous driving strategies [23]. For instance, Huang *et al.* [24] leveraged human prior knowledge to tackle the challenges of sample efficiency and reward function design in DRL. The effectiveness of their method was validated in unprotected left-turn and roundabout scenarios. Li *et al.* [25] proposed a hierarchical skill-based offline reinforcement learning method to solve the long-horizon vehicle planning task. Dang *et al.* [26] developed an event-triggered model predictive control framework based on a model-free DRL algorithm to address the path following problem in autonomous driving. Although the above DRL-based driving policies have achieved remarkable advancements in many scenarios, the lack of robustness against adversarial attacks limits their application in real-world settings [27].

*B. Attack Methods for DRL-based Autonomous Driving*

Existing studies have demonstrated that well-trained DRL policies remain highly vulnerable when facing adversarial perturbations [17]-[20]. Adding adversarial perturbations to state inputs in autonomous driving policies can cause sudden acceleration or braking of the ego vehicle, posing a safety risk. Consequently, some studies have attempted to explore the robustness of autonomous driving policies under adversarial environments and have proposed some adversarial attack methods. These attack methods can be broadly categorized into two types: untargeted attacks [28]-[30] and targeted attacks [14][31][32].

Untargeted attacks typically aim to maximize the distance between the agent's policies before and after the perturbation. For instance, He *et al.* [33] proposed a black-box attack technique with Bayesian optimization to maximize the action deviation of the ego vehicle by introducing adversarial perturbations to its observations. However, these methods may fail to induce worst-case scenarios, such as collisions, due to the lack of a clear purpose. Therefore, some targeted attack methods are proposed with the aim of minimizing the agent's reward or maximizing safety violations. For instance, Ma *et al.* [34] introduced an adversary that can introduce up to 20% adversarial perturbations to the actions of the autonomous driving agent to minimize the agent's reward.

All the aforementioned attack methods assume that adversarial perturbations are continuously applied throughout the driving process. However, excessive perturbations make the attack more detectable and easier to mitigate. Furthermore, continuous attacks may fail to significantly improve the effectiveness, as the critical moments for attacks are sparse. Non-critical perturbation samples may also interfere with the adversarial training of the autonomous driving agent, hindering its ability to improve robustness. We believe that considering the sparse distribution of critical attack moments in autonomous driving is essential. Based on this insight, we propose a novel stealthy and efficient adversarial attack method.

III. PROBLEM FORMULATION

This section first describes the preliminary concept of MDP in Section III.A. Then we model the attack problem as a mathematical formulation according to the goal of the proposed attack in Section III.B. Section III.C defines the threat model.

*A. Preliminary*

Driving policies can be learned through DRL algorithms [1]. Before learning the policy, it is usually necessary to model the driving policy learning problem as an MDP which can be defined by a tuple $(\mathcal{S}, \mathcal{A}, r, \mathcal{P}, \gamma)$. Here state space $\mathcal{S}$ denotes the set of all possible states of the system. In autonomous driving, the state information typically includes information about the ego vehicle, surrounding vehicles, and driving environment. Action space $\mathcal{A}$ represents the set of all possible control actions such as steering and throttle. The reward function $r: \mathcal{S} \times \mathcal{A} \to \mathbb{R}$ is designed to guide the agent to learn the desired driving policy and $\mathcal{P}: \mathcal{S} \times \mathcal{A} \times \mathcal{S} \to [0,1]$ describes the transition probability of the system. $\gamma \in (0,1)$ is the discount factor. The DRL-based agent aims to find an optimal policy $\pi^*$ that maximizes the cumulative discounted reward

$$J(\pi) = \mathbb{E}_{a_t \sim \pi(s_t)} \left( \sum_{t=1}^{T} \gamma^t r(s_t, a_t) \right). \tag{1}$$

*B. Problem Statement*

In this paper, we aim to find a stealthy and efficient attack method against DRL-based autonomous driving policies. Instead of attacking at every time step, a more efficient and stealthy approach would be to limit the attacks to a few critical time steps while still achieving the desired objective. Let $\bar{r}$ as the reward function of the adversary, we can then formulate the above intuition as the following optimization problem:

$$\max_{(x,\delta)} \mathbb{E}_{a'_t \sim \pi(s'_t)} \left( \sum_{t=1}^{T} \gamma^t \bar{r}(s'_t, a'_t) \right), \tag{2}$$

s.t.

$$s'_t = s_t + x_t \delta_t, t \in T, \tag{2a}$$

$$\delta_t = PG(s_t, a_t, a'_t), t \in T, \tag{2b}$$

$$\delta_t \leq \epsilon, t \in T, \tag{2c}$$

$$\sum_{t \in T} x_t \leq \Gamma. \tag{2d}$$

Eq. (2) is the objective function that maximizes the adversary's cumulative discounted reward over the entire episode, where $T$ is the maximum length of each episode. The variables are denoted as $(x, \delta)$, where $x = \{x_t, t \in T\}$ is a set of binary variables used to control whether to apply the attack at each time step, and $\delta = \{\delta_t, t \in T\}$ denotes the set of perturbations. If $x_t = 1$, the adversary will add perturbations $\delta_t$ to the state $s_t$, as described in (2a). $\delta_t$ can be generated by perturbation generating method $PG(\cdot)$. As shown in (2c), we set a threshold $\epsilon$ for $\delta_t$ to further reduce the probability of the attack being detected. We assume the total number of attacks per episode must not exceed $\Gamma$, as shown in (2d), where $\Gamma \ll T$. Solving this optimization problem is challenging because it is a complex mixed integer optimization problem. The difficulty is further complicated by the non-linear growth in the size of the problem, driven by $T$ [17].

*C. Threat Model*

We outline the goal and capabilities of the adversary to guide the design of the attack policy before detailing our attack method to solve the optimization problem (2).

*Goal of the adversary*: The adversary attempts to manipulate the agent's actions by introducing small adversarial perturbations to its observations at critical time steps, thereby causing safety violations such as collisions.

*Capabilities of the adversary:* We assume that the adversary can only access the observations and actions of the agent. Furthermore, the adversary is only permitted to modify the agent's observations by adding adversarial perturbations [35]. Since the adversary has no knowledge of the architecture, parameters, or gradients of the agent's policy, our attack is a black-box attack.

## IV. METHODOLOGY

This section first outlines the overall framework of our attack method in Section IV.A and propose the algorithm in Section IV.B.

*A. Framework*

The optimization problem can be easily divided into two subproblems: **when-to-attack** and **how-to-attack**. In our attack method, we introduce a DRL-based adversary to independently learn the optimal strategies for both subproblems. The overview diagram of the proposed attack method is given in Fig. 2. Our attack method consists of two stages. The first stage is the decision-making stage. In this stage, the adversary needs to make attack decisions $(p_t, a'_t) \sim \pi_{adv}(s_t^{adv})$ based on its observation $s_t^{adv}$. Besides treating $s_t^{adv} = s_t$ as in [19], we believe that introducing additional attack-related information can enhance the ability of the adversary. Therefore, we add the remaining number of attacks $n_t$ and the agent's original action $a_t$ into the adversary's observation, i.e., $s_t^{adv} = (s_t, n_t, a_t)$. The adversary's attack decision consists of two parts, where $p_t \in [-1,1]$ represents the switch action determining whether to inject the perturbation into $s_t$ to launch an attack, and $a'_t \in \mathcal{A}$ represents the lure action that the adversary aims to induce the agent to take by applying the perturbation.

After making the attack decision, the adversary enters the next stage, the attack launching stage. In this stage, the adversary will make the final decision based on $(p_t, a'_t)$. If $p_t \geq 0$, the adversary will generate the perturbation $\delta_t$ to mislead the agent to trigger the action $a'_t$. Conversely, if $p_t < 0$, it indicates that no attack will be launched at time step $t$. After this stage, the adversary will terminate the episode if $n_t = 0$, this process also called trajectory clipping. The purpose is to balance the sample distribution and enhance the training effectiveness of the adversary.

Our attack method is capable of learning optimal switch and lure policies automatically without any domain knowledge. The domain-agnostic and self-learning characteristics make our method highly suitable for complex and diverse autonomous driving scenarios.

*B. Algorithm*

Given that our autonomous driving scenario involves a continuous action space, we choose PPO, a state-of-the-art on-policy DRL algorithm [36], to train the adversary. PPO consists of two networks, an actor network $\pi_\theta^{adv}$ and a critic network $V_\phi^{adv}$, where $\theta$ and $\phi$ are the parameters corresponding to these two networks. Given an adversary observation $s_t^{adv}$, $\pi_\theta^{adv}$ will output the policy $\pi_\theta^{adv}((p_t, a'_t) | s_t^{adv})$. and $V_\phi^{adv}$ will estimate the value of $s_t^{adv}$, which can be expressed as

$$V_\phi^{adv}(s_t^{adv}) = \mathbb{E}\left[\sum_{k=1}^{T-t} \gamma^{k-1} \overline{r}_{t+k} \mid s_t^{adv}\right]. \quad (3)$$

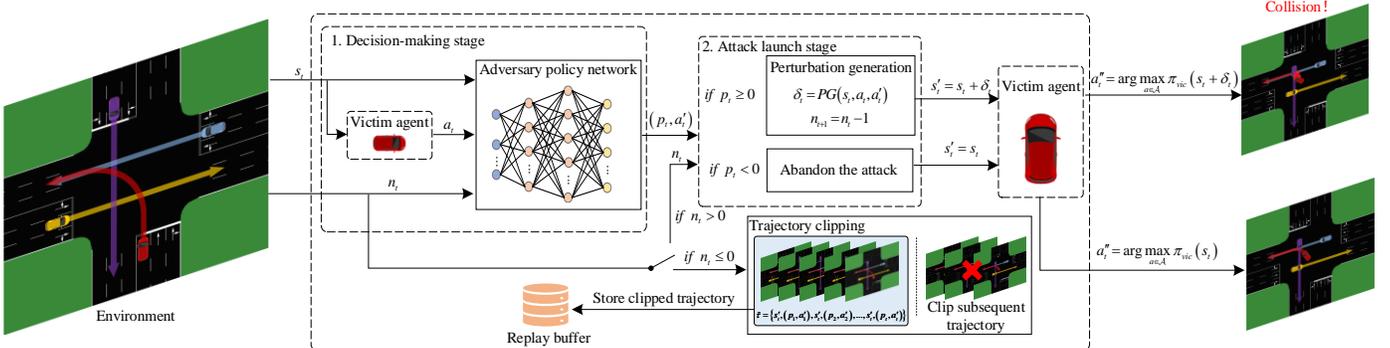

**Fig. 2.** The framework of our attack method.

The loss function used to optimize $\pi_\theta^{adv}$ can be expressed as

$$L_a(\theta) = \mathbb{E}_t\left[\min\left(\rho_t^{clip}(\theta), \rho_t(\theta)\right) \hat{A}_t\right], \quad (4)$$

where

$$\rho_t^{clip}(\theta) = clip(\rho_t(\theta), 1-\varepsilon, 1+\varepsilon), \quad (5)$$

$$\rho_t(\theta) = \frac{\pi_\theta^{adv}((p_t, a_t') | s_t^{adv})}{\pi_\theta^{adv,old}((p_t, a_t') | s_t^{adv})}, \quad (6)$$

where $\rho_t(\theta)$ represents the probability ratio between the current policy $\pi_\theta^{adv}$ and the old policy $\pi_\theta^{adv,old}$. $\hat{A}_t$ is the estimated advantage and can be defined as

$$\hat{A}_t = \sum_{i=t}^{T-1}(\gamma\lambda)^{i-t}(\overline{r}_i + \gamma V_\phi^{adv}(s_{t+1}^{adv}) - \gamma V_\phi^{adv}(s_t^{adv})), \quad (7)$$

where $\lambda$ is a tuning parameter.

The loss function for the critic network is typically a mean squared error, which can be expressed as

$$L_c(\phi) = \mathbb{E}_t\left[\left(V_\phi^{adv}(s_t^{adv}) - \left(\overline{r}_i + \gamma V_\phi^{adv}(s_{t+1}^{adv})\right)\right)^2\right]. \quad (8)$$

Then we can obtain the total loss function as

$$L(\theta, \phi) = L_a(\theta) - c_1 L_c(\phi) + c_2 S\left[\pi_\theta^{adv}\right](s_t^{adv}), \quad (9)$$

where $c_1$ and $c_2$ are coefficients and $S[\cdot]$ denotes an entropy bonus. Based on the loss function, $\theta$ and $\phi$ can be simultaneously updated using gradient-based methods.

---

**Algorithm 1 Our attack method**

Input: Environment $\mathcal{E}$, victim agent policy $\pi_{vic}$
Output: The optimal adversarial policy network $\pi_{adv}^*$
Initialize network parameters $\theta$ and $\phi$, experience replay buffer $D$
1:  **For** $episode = 1, 2, ..., L$ **do**:
2:      $n_1 = \Gamma$
4:      **For** $t = 1, 2, ..., T$ **do**:
5:          $a_t = \arg\max_{a \in \mathcal{A}} \pi_{vic}(a | s_t)$, $(p_t, a_t') \sim \pi_{adv}(s_t, a_t, n_t)$
6:          **If** $p_t > 0$ **and** $n_{t-1} > 0$ **then**
7:              $n_{t+1} = n_t - 1$
8:              $\delta = PerturbationGenerate(s_t, a_t, a_t')$
9:              $s_t' = s_t + \delta$
10:         **Else**
11:             $s_t' = s_t$, $n_{t+1} = n_t$
12:         $a_t'' = \pi_{vic}(s_t')$
13:         Perform $a_t''$ and obtain $(s_{t+1}, \overline{r}_t)$
14:         Store $\tau_t = (s_t, (p_t, a_t'), s_{t+1}, \overline{r}_t)$ into $D$
15:         **If** $n_{t+1} = 0$ **then**
16:             Clip the trajectory and reset the environment $\mathcal{E}$
19:     **End for**
20:     Sample from $D$ to update $\theta$ and $\phi$ according to (9)
21: **End for**

---

A step-by-step outline of our attack method is given in Algorithm 1. First, initialize the network parameters $\theta$ and $\phi$, as well as the experience replay buffer $D$. Next, the adversary will interact with the environment for $L$ episodes, with each episode having a maximum length of $T$. At each time step, the adversary generates $(p_t, a_t')$ based on its observations. Then the adversary chooses to apply perturbations to the agent's observations at critical time steps, and stores the samples into the buffer. Finally, the adversary samples data from the buffer and updates the policy parameters according to (9).

## V. EXPERIMENT

We validate the proposed attack method on the SUMO platform. First, we introduce the relevant experimental settings in Sections V.A-E, including environment settings, the MDP, baselines, training details, and metrics. Then we organize our experiments and explain the results to address the following key questions.

• **RQ1**: Can the proposed attack method can achieve the adversarial goal within a limited number of attacks? (Studied in Section V.F1)

• **RQ2**: Does the proposed attack method outperform the baseline and SOTA methods? (Studied in Section V.F2)

• **RQ3**: How does the value of $\Gamma$ affect the proposed attack method? (Studied in Section V.F3)

• **RQ4**: How does the performance of our proposed attack method vary under different traffic densities? (Studied in Section V.F4)

### A. Environment Setting

We select an unprotected left-turn scenario for experiment, where the ego vehicle needs to complete its left turn while interacting with dynamic cross-traffic flows at an unsignalized intersection, as shown in Fig. 2. In the simulated environment, all vehicles employ the LC2013 model of SUMO for lane-changing behavior. To ensure realistic dynamics, the maximum speed $v_{max}$ is set to 15 m/s and maximum acceleration (and deceleration) $\beta$ of the ego vehicle are set to 15 m/s and 7.6 m/s², respectively. The traffic flow density is determined by the probability $p$ of vehicle arrivals per second. If there is no special explanation, we set $p = 0.5$.

### B. Markov Decision Process

Based on the simulation environment, we construct an MDP with the following details:

**State Space** $\mathcal{S}$: The state $s_t \in \mathcal{S}$ consists of two components. One is the speed and velocity direction of the ego vehicle. The other part contains information about the six nearest vehicles within a 200-meter radius of the ego vehicle, including the relative distance, orientation, speed, and velocity direction. The positions of these six cars are front, rear, left front, left rear, right front and right rear.

**Action Space** $\mathcal{A}$: At time step $t$, the action of the agent $a_t$ involves continuous acceleration or deceleration, represented as a continuous variable within the range [-1,1]. To accurately reflect real conditions, $a_t$ needs to be multiplied by the acceleration coefficient $\beta$ to obtain the actual acceleration of the ego vehicle. Thus, the velocity update equation can be obtained as $v_{ego}(t+1) = v_{ego}(t) + a_t \beta \Delta t$.

**Reward function** $r$: The reward function of the victim agent $r_t$ consists of two components: efficiency reward and safety penalty, which can be expressed as:

$$r_t = \frac{v_t}{v_{max}} - c(s_t, a_t''),$$
$$c(s_t, a_t'') = \begin{cases} 1, & \text{if collision,} \\ 0, & \text{else,} \end{cases} \quad (10)$$

where $v_{max}$ denotes the maximum speed of the ego vehicle and $a_t''$ represents the action performed by the ego vehicle at time step $t$. Specifically, the goal of the victim agent is to drive the ego vehicle as quickly as possible while avoiding collisions. Since the adversary aims to cause a collision, setting its reward $\bar{r}_t = -r_t$ as in [20] is unreasonable. Because it may redirect the adversary's focus towards reducing the speed of the ego vehicle. Therefore, we set $\bar{r} = c(s_t, a_t'')$ so that the adversary focuses on causing safety violations.

*C. Baselines*

We select autonomous driving agents based on state-of-the-art RL algorithms to verify the effectiveness of our attack method. These agents can be divided into two categories: one is based on vanilla RL algorithms, such as PPO and the state-of-the-art off-policy algorithms SAC [37] and TD3 [38]. The other is the state-of-the-art safe autonomous driving DRL algorithm, FNI-RL [39].

In terms of attack methods, we selected the following attack methods for comparison with our proposed attack method.

*Random Attack (RA)*: At each time step, the adversary randomly samples an adversarial action from the action space. Then perturbations are generated to guide the agent toward selecting an action aligned with the adversarial one. Since adversarial actions are randomly selected, RA is a typical non-targeted attack method.

*Action Modification Attack (AMA)*: In this attack, the adversary possesses nearly the same capabilities as our attack method. The difference is that this attack directly forces the adversarial action as the final executed action, rather than guiding the agent through perturbations, i.e., $a_t'' = a_t'$. It is typically unrealistic to directly manipulate the agent's action in practical applications, as it requires the adversary to have higher privileges. However, this attack method can serve as a strong baseline to verify the effectiveness of our attack. Specifically, the effect of this attack is intuitively not weaker than our attack method since each attack can achieve the expected target with a 100% success rate.

*Unlimited Attack (UA)*: This attack is similar to the attack methods commonly adopted in current work, as it performs an attack at each time step. This attack generates the adversarial action in the same way as our method.

*D. Training details*

To verify the effectiveness of our attack method, we first train several victim agents using different DRL algorithms. All agents were trained for 12,000 steps, with a maximum episode length $T$ set to 30. All DRL algorithms are implemented using Stable Baselines3 [40]. We fine-tune the hyperparameters of the PPO-based victim agent for better performance. Specifically, the batch size is set to 256, the learning rate to 0.001, and the number of epochs to 15. The parameters for the other agents are set to the default values provided by Stable Baselines3. Except for the random attack, all other attack methods are PPO-based. Therefore, we train the adversaries based on the victim agents to generate the corresponding attack policies. The total training steps and maximum episode length are consistent with the victim agents, and the hyperparameters, such as learning rate and batch size are set to default values. We select FGSM [41] and PGD [42], two generalized adversarial sample generation approaches, to create adversarial perturbations. All the experiments are conducted on a server equipped with Intel(R) Xeon(R) Gold 6230 CPUs and NVIDIA GeForce RTX 4090 GPUs.

*E. Metrics*

We select multiple metrics to conduct a comprehensive evaluation of our attack method. Besides the common metrics such as success rate (SR), collision rate (CR), average speed (AS), and average reward (AR) [39], we also adopt the average number of attacks (ANA) to evaluate the efficiency of our attack method. Specifically, ANA refers to the number of attacks launched by the adversary in an episode. Moreover, we introduce a new metric to evaluate the effectiveness of a single attack, called attack efficiency (AE). AE can be calculated using CR and ANA, as follows:

$$AE = CR \cdot e^{-k \cdot ANA}, \quad (11)$$

where $k$ is a parameter used to adjust the weights of CR and ANA. In this paper, it is set to 0.05. All metrics are computed based on the results of 100 test episodes.

*F. Performance Evaluation*

In this section, we aim to address RQ1-RQ4 through experiments and rigorous analysis.

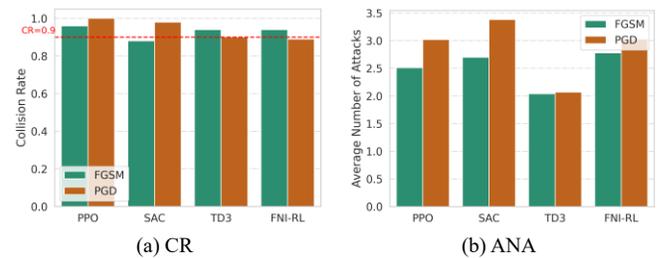

(a) CR  (b) ANA
**Fig. 3.** The performance of our attack method against different victim agents.

*F1. Effectiveness of Our Method*

In this section, we will verify that our approach enables stealthy and efficient attacks, i.e., the highest possible collision rate with as few attacks as possible. Fig. 3 illustrates the performance of our method against victim agents trained by PPO, SAC, TD3, and FNI-RL. It can be observed that, in most cases, our attack method achieves a collision rate of around 90% within an average of three attacks per episode. Both FGSM-based and PGD-based attacks achieve similar attack effectiveness, i.e., CR.

TABLE I.

STATISTICAL RESULTS OF AUTONOMOUS DRIVING AGENTS UNDER DIFFERENT ADVERSARIAL ATTACK METHODS

| Method | Metrics | No Attack | AMA | RA | | UA | | Ours | |
|---|---|---|---|---|---|---|---|---|---|
| | | | | FGSM | PGD | FGSM | PGD | FGSM | PGD |
| PPO | SR | 0.90 | **0.00** | 0.47 | 0.61 | 0.02 | 0.05 | 0.04 | 0.04 |
| | CR | 0.10 | **1.00** | 0.46 | 0.39 | 0.98 | 0.95 | 0.96 | 0.96 |
| | AS | 13.09±3.88 | 13.66±3.86 | 9.80±4.65 | 11.52±4.26 | 13.59±3.78 | 13.60±3.72 | 13.46±3.89 | 13.48±3.91 |
| | AR | 25.69±4.42 | 14.23±2.80 | 16.90±4.56 | 19.88±5.82 | 14.82±3.79 | 15.79±4.48 | 15.65±4.38 | 15.15±4.08 |
| | ANA | - | **1.59±1.19** | 25.77±4.94 | 25.62±5.64 | 17.42±3.69 | 18.40±4.41 | 2.51±1.15 | 2.66±1.56 |
| | AE | - | **0.92** | 0.13 | 0.11 | 0.41 | 0.38 | 0.85 | 0.84 |
| SAC | SR | 1.00 | **0.00** | 0.52 | 0.60 | 0.06 | 0.21 | 0.12 | 0.02 |
| | CR | 0.00 | **1.00** | 0.47 | 0.39 | 0.94 | 0.79 | 0.88 | 0.98 |
| | AS | 13.21±3.36 | 13.47±3.90 | 10.55±4.55 | 11.02±4.39 | 13.33±3.80 | 13.43±3.61 | 13.53±3.80 | 13.60±3.85 |
| | AR | 27.31±0.61 | 15.12±4.26 | 17.21±4.78 | 19.09±5.30 | 17.03±4.43 | 19.83±6.10 | 15.25±4.99 | 14.44±3.50 |
| | ANA | - | 3.90±1.92 | 24.58±6.15 | 25.75±5.60 | 20.12±4.26 | 22.78±5.79 | **2.70±0.83** | 3.38±2.03 |
| | AE | - | 0.82 | 0.14 | 0.11 | 0.34 | 0.25 | 0.77 | **0.83** |
| TD3 | SR | 0.72 | 0.06 | 0.19 | 0.24 | **0.00** | 0.01 | 0.06 | 0.10 |
| | CR | 0.20 | 0.94 | 0.76 | 0.73 | **1.00** | 0.99 | 0.94 | 0.90 |
| | AS | 12.73±4.48 | 13.35±3.94 | 12.00±4.78 | 11.80±4.90 | 13.51±3.86 | 13.48±3.84 | 16.61±5.10 | 13.19±4.03 |
| | AR | 25.58±4.60 | 16.55±4.90 | 17.26±5.68 | 17.59±5.42 | 14.72±4.06 | 15.06±4.14 | 13.22±3.99 | 16.84±5.12 |
| | ANA | - | **1.92±0.39** | 22.21±5.72 | 22.93±5.73 | 17.42±4.10 | 17.83±4.09 | 2.04±0.53 | 2.07±0.75 |
| | AE | - | **0.85** | 0.25 | 0.23 | 0.42 | 0.41 | **0.85** | 0.81 |
| FNI-RL | SR | 0.98 | 0.06 | 0.45 | 0.60 | **0.00** | 0.03 | 0.06 | 0.11 |
| | CR | 0.02 | 0.94 | 0.49 | 0.38 | **1.00** | 0.97 | 0.94 | 0.89 |
| | AS | 12.72±3.80 | 13.47±3.93 | 10.92±4.91 | 10.65±5.05 | 13.66±3.86 | 13.63±3.88 | 13.43±3.96 | 13.45±3.91 |
| | AR | 26.05±2.07 | 14.59±3.81 | 17.29±5.04 | 18.51±4.91 | 14.23±2.80 | 14.56±3.78 | 14.54±3.84 | 14.97±4.69 |
| | ANA | - | 2.82±2.02 | 23.89±6.42 | 26.00±5.45 | 16.72±2.80 | 17.06±3.66 | **2.78±2.22** | 3.03±2.15 |
| | AE | - | 0.82 | 0.15 | 0.10 | 0.43 | 0.41 | **0.82** | 0.76 |

*F2. Performance Comparison*

To further evaluate the effectiveness of our method, we compare the performance of our method with four baseline approaches (including the scenario without attack). The results are shown in Table I. It can be observed that all attacks reduce the reward of the ego vehicle. Because the attacks increase the likelihood of collisions, these collisions subsequently prevent the ego vehicle from receiving future rewards. Moreover, except for RA, all other attacks lead to an increase in the average speed of the ego vehicle compared to the no-attack case. Because the other attacks are targeted attacks, which aim specifically to cause the ego vehicle to collide. Such collisions are often preceded by higher speeds. Compared to AMA and UA, our method also achieves optimal or near-optimal performance in CR. In contrast to our method and AMA, the effectiveness of UA relies on the continuous addition of perturbation. Consequently, UA lacks both stealthy and efficiency. Leveraging the precision of action modification, AMA achieves a 100% collision rate in most cases. However, it is important to note that directly modifying actions may be more easily detectable compared to UA. The purpose of choosing AMA as a baseline is to make it an ideal case for our attack method. Finally, we compared our attack method with other algorithms to validate its attack efficiency. Compared to UA and RA, our method achieves an average improvement of 160.47% and 654.48% in AE, respectively.

*F3. Impact of the Upper Bound of Attack Steps*

Since we expect the adversary to trigger collisions through a limited number of attack steps, we control the upper bound on the number of attack steps per episode during the training process. Fig. 4 shows the performance of our attack method under different $\Gamma$. The victim agent is based on FNI-RL, and the perturbation generation method is FGSM. The vertical axis represents the total reward of the adversary. Notably, this reward is binary, taking a value of 1 for collisions and 0 for non-collisions. Therefore, convolutional smoothing is applied to the reward curves for better representation. After smoothing, the vertical axis can be interpreted as a proxy for the collision probability to some extent. It can be observed that as $\Gamma$ increases, the collision probability increases, indicating an improvement in the attack capability of the adversary. This is because a larger upper bound enables the adversary to explore more situations, thereby increasing the likelihood of learning the optimal attack policy. Similar conclusions can be drawn in other cases where the victim agent and perturbation generation method are different.

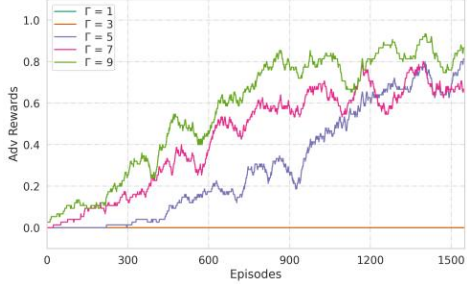

**Fig. 4.** Training reward curves under different upper bounds of attack steps.

It is noteworthy that there is no significant positive correlation between $\Gamma$ and the actual number of attack steps performed by the adversary. To prove this, we test the attack policies obtained under different $\Gamma$ with 100 episodes, the results are shown in Table II. It can be observed that the average number of attack steps launched by the attack policies is far from reaching $\Gamma$, remaining around 3 times per episode.

TABLE II.

CR, AND ANA OF ATTACK POLICIES UNDER DIFFERENT $\Gamma$

|  | CR | ANA |  | CR | ANA |
|---|---|---|---|---|---|
| $\Gamma = 5$ | 0.86 | 2.58±1.25 | $\Gamma = 8$ | 0.88 | 3.27±2.30 |
| $\Gamma = 6$ | 0.86 | 3.24±1.23 | $\Gamma = 9$ | 0.94 | 3.10±2.05 |
| $\Gamma = 7$ | 0.86 | 3.58±1.92 | $\Gamma = 10$ | 0.94 | 3.00±2.30 |

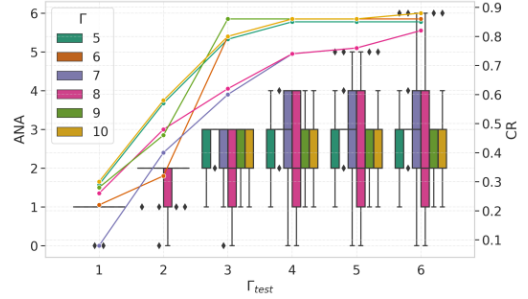

**Fig. 5.** Performance under different $\Gamma_{test}$.

TABLE III.

STATISTICAL RESULTS OF ATTACK METHODS UNDER DIFFERENT TRAFFIC DENSITIES

| Method | PG | Metrics | PPO | | SAC | | TD3 | | FNI-RL | |
|---|---|---|---|---|---|---|---|---|---|---|
| | | | Flow-1 | Flow-2 | Flow-1 | Flow-2 | Flow-1 | Flow-2 | Flow-1 | Flow-2 |
| No-attack | - | CR | 0.32 | 0.06 | 0.12 | 0.06 | 0.20 | 0.16 | 0.02 | 0.00 |
| | | ANA | - | - | - | - | - | - | - | - |
| | | AE | - | - | - | - | - | - | - | - |
| AMA | - | CR | 0.78 | 0.92 | 0.76 | 0.88 | **0.84** | **0.98** | 0.70 | 0.96 |
| | | ANA | **2.34±1.54** | 2.80±1.31 | 2.92±1.38 | 2.90±0.67 | 2.74±0.89 | **1.82±0.38** | **3.18±2.95** | 2.96±2.07 |
| | | AE | 0.69 | 0.80 | 0.66 | 0.76 | **0.73** | **0.89** | **0.60** | **0.83** |
| RA | FGSM | CR | 0.28 | 0.55 | 0.48 | 0.47 | 0.69 | 0.81 | 0.42 | 0.39 |
| | | ANA | 27.66±4.17 | 25.16±5.16 | 24.61±6.09 | 25.02±5.77 | 22.98±5.92 | 22.20±5.50 | 25.33±5.89 | 25.27±6.08 |
| | | AE | 0.07 | 0.16 | 0.14 | 0.13 | 0.22 | 0.27 | 0.12 | 0.11 |
| | PGD | CR | 0.35 | 0.42 | 0.43 | 0.51 | 0.60 | 0.75 | 0.38 | 0.42 |
| | | ANA | 26.11±5.48 | 25.43±5.57 | 24.83±6.25 | 24.32±5.98 | 23.92±5.94 | 23.74±5.21 | 25.62±6.04 | 25.12±5.91 |
| | | AE | 0.09 | 0.12 | 0.12 | 0.15 | 0.18 | 0.23 | 0.11 | 0.12 |
| UA | FGSM | CR | 0.70 | 0.74 | **0.84** | **1.00** | 0.84 | **0.98** | 0.80 | **1.00** |
| | | ANA | 22.66±5.69 | 22.08±5.52 | 18.98±5.64 | 17.50±3.65 | 19.40±5.69 | 17.30±3.52 | 19.70±6.07 | 16.74±2.79 |
| | | AE | 0.25 | 0.41 | 0.33 | 0.41 | 0.32 | 0.41 | 0.30 | 0.43 |
| | PGD | CR | **0.85** | **1.00** | **0.84** | 0.97 | 0.84 | 0.97 | 0.78 | 0.97 |
| | | ANA | 18.79±5.65 | 16.76±2.85 | 19.06±5.64 | 17.75±3.92 | 19.54±5.74 | 17.61±3.47 | 19.65±6.22 | 17.07±3.49 |
| | | AE | 0.43 | 0.40 | 0.32 | 0.40 | 0.32 | 0.40 | 0.29 | 0.41 |
| Ours | FGSM | CR | 0.78 | 0.92 | 0.74 | 0.88 | 0.82 | 0.92 | 0.64 | 0.96 |
| | | ANA | 2.43±1.52 | **2.80±1.26** | **2.76±1.21** | **2.80±0.63** | 3.00±1.22 | 1.94±0.42 | 3.62±3.39 | 3.16±2.45 |
| | | AE | 0.69 | 0.80 | 0.64 | 0.77 | 0.71 | 0.83 | 0.53 | 0.82 |
| | PGD | CR | 0.84 | 0.98 | 0.80 | 0.96 | 0.75 | 0.93 | 0.66 | 0.88 |
| | | ANA | 3.12±2.21 | 3.32±1.59 | 3.51±2.29 | 3.38±2.08 | **2.08±0.90** | 1.84±0.42 | 3.67±2.64 | **2.84±1.80** |
| | | AE | **0.72** | **0.83** | **0.67** | **0.81** | 0.68 | 0.85 | 0.55 | 0.76 |

We then adjust the upper bound $\Gamma_{test}$ on the number of attack steps during testing to evaluate the performance of our method, as shown in Fig. 5. It can be observed that as $\Gamma_{test}$ increases, the attack policies trained under different $\Gamma$ are able to achieve progressively higher collision rates. The reason is that perturbation generation methods cannot guarantee a 100% success rate. Therefore, it is necessary to launch subsequent attacks based on the outcome of previous attacks to mitigate the biases introduced by the perturbation generation method. Nevertheless, our method achieves more than 70% CR within 3 to 4 attack steps in most cases. This indicates that our attack method can achieve significant effectiveness by launching attacks in about 10% of the time steps within an episode, effectively balancing stealthy and efficiency.

*F4. Performance Under Different Traffic Densities*

To conduct a more comprehensive evaluation, we construct two new scenarios with different traffic densities, denoted as flow1 and flow2. Specifically, flow1 and flow2 represent the cases where $p=0.3$ and $p=0.7$ respectively. Table III shows the performance of the attack methods under flow1 and flow2 scenarios. It can be observed that as traffic flow density increases, all attack methods can increase the collision rate in most cases. Similar to the results in Table II, our method consistently achieves comparable performance to AMA in all cases. This implies that existing perturbation generation methods can already yield satisfactory results in our scenarios. Compared to UA and RA, our attack method achieves highly competitive collision rates in all cases with significantly fewer attack steps. Specifically, compared to UA and RA, our attack method achieves an average improvement of 141.24% and 607.23% in AE in flow1, respectively. In flow2, the improvements are 130.65% and 596.62%, respectively.

## VI. CONCLUSION

In this work, we propose a novel adversarial attack method that enables stealthy and efficient attacks against DRL-based autonomous driving agents. The proposed attack method is based on DRL, allowing it to learn the optimal attack strategy in a domain-independent and self-learning manner. We validate the effectiveness of our attack across various state-of-the-art DRL algorithms, including both vanilla and safe DRL algorithms. Moreover, we compare our method with existing attack approaches. Experimental results show that our attack method achieves over 90% collision rate in most cases while the attack efficiency is significantly higher than other attack methods. Our attack method also shows robustness to changes in traffic density. Although we have demonstrated the potential of our attack method, there remain some limitations that should be addressed in future studies. Firstly, developing more efficient perturbation generation methods is expected to further enhance the effectiveness of our attack method. Secondly, adapting our attack method to long-term autonomous driving scenarios presents an interesting area for further research. Finally, investigating effective defense strategies against our attack method will contribute to improving the robustness of autonomous driving systems.